\documentclass[runningheads]{llncs}

\usepackage{eccv}

\usepackage{eccvabbrv}

\usepackage{graphicx}
\usepackage{booktabs}
\usepackage{enumitem}
\usepackage{multirow}
\usepackage{colortbl}

\usepackage[accsupp]{
  axessibility
} 

\usepackage{hyperref}
\usepackage[capitalize]{cleveref}

\usepackage{orcidlink}

\begin{document}
\title{ReactVAU: A Slow-Fast Decoupled Framework for Streaming Video Anomaly Understanding}

\titlerunning{ReactVAU: Streaming Video Anomaly Understanding}

\author{Chia-Hui Chen\inst{1}\orcidlink{0009-0001-7469-5031} \and Shih-Ying
  Yeh\inst{1}\orcidlink{0009-0002-5962-091X} \and Fu-En Yang\inst{2}\orcidlink{0000-0003-0102-7101}
  \and Min-Hung Chen\inst{2}\orcidlink{0000-0002-4046-3937} \and Shang-Hong Lai\inst{1}\orcidlink{0000-0002-5092-993X}
}

\authorrunning{C.-H.~Chen et al.}

\institute{National Tsing Hua University, Hsinchu, Taiwan \and NVIDIA, Taipei, Taiwan}

\maketitle

\begin{abstract}
  In this paper, we propose \textbf{ReactVAU}, a Slow-Fast Decoupled Framework
  for real-time streaming Video Anomaly Understanding (VAU). Existing VAU methods
  rely on offline inference with global temporal sampling, which violates
  causality and prevents deployment in live surveillance streams. Conversely,
  general streaming video models satisfy causal access but dilute rare
  transient anomalies during memory compression and often invoke heavyweight
  MLLMs uniformly over long normal intervals. ReactVAU addresses this gap with
  three synergistic components: a lightweight Fast Detection Module based on
  Spatial Grid Folding (SGF) for continuous anomaly filtering; an Anomaly-Aware
  Persistent Memory (AAPM) that protects critical visual cues from temporal decay;
  and a heavyweight Slow Reasoning Module that remains dormant during normal
  streams and is awakened only by suspicious events for semantic verification
  and causal description. Extensive experiments on multiple benchmarks demonstrate
  that ReactVAU operates under strict streaming constraints while simultaneously
  achieving competitive performance in both anomaly detection and causal reasoning,
  alongside significantly enhanced computational efficiency by minimizing
  heavyweight MLLM invocations. Project page is available at \href{https://huiyuiui.github.io/ReactVAU/}{https://huiyuiui.github.io/ReactVAU/}

\end{abstract}

\section{Introduction}
\label{sec:intro} Video anomaly detection (VAD) serves as a foundational
technology for intelligent surveillance, traffic monitoring, and public safety.
Early VAD methodologies inherently operated at the frame level to identify deviations
from normal patterns \cite{lu2013abnormal, hasan2016learning, tran2015learning,
liu2018future, sultani2018real, gong2019memorizing, wu2020not}. While
effective for localization, they functioned largely as black boxes, providing
only a scalar anomaly probability without semantic context regarding what
occurred or why it was deemed anomalous. To meet the real-world demand for semantic
explainability, recent advancements introduced Vision-Language Models (VLMs)
and Multimodal Large Language Models (MLLMs)
\cite{wu2023vadclipadaptingvisionlanguagemodels, zanella2024harnessinglargelanguagemodels,
  zhang2024holmesvadunbiasedexplainablevideo, ahn2025anyanomalyzeroshotcustomizablevideo,
chen2025aligningeffectivetokensvideo}, driving a significant paradigm shift from
traditional VAD to comprehensive Video Anomaly Understanding (VAU). However, current
top-tier VAU models are inherently constrained by offline inference paradigms \cite{du2024uncoveringwhathowcomprehensive,
  zhang2025holmesvaulongtermvideoanomaly, ye2025veraexplainablevideoanomaly, huang2025vadr1videoanomalyreasoning,
zhu2025vaur1advancingvideoanomaly, lin2025unifiedreasoningframeworkholistic}.
These architectures rely on observing the entire video sequence to perform
global temporal sampling or calculate holistic anomaly density distributions
prior to language generation. This absolute dependence on future frames breaks
causality, rendering them fundamentally incompatible with real-world streaming
surveillance environments.

\begin{figure*}[t]
  \centering
  \includegraphics[width=\linewidth]{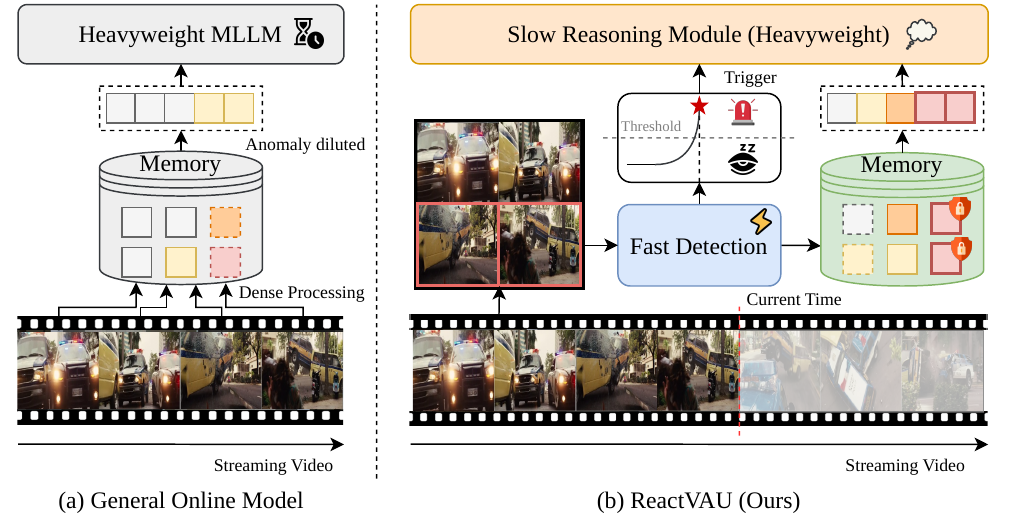}
  \caption{\textbf{Conceptual comparison of ReactVAU against general online
    models.} (a) General Online Model: Continuously evaluating dense video
    streams with heavyweight MLLMs incurs prohibitive computational overhead, and
    anomaly features are easily diluted during memory compression. (b) ReactVAU:
    A lightweight Fast Module continuously monitors streams and reactively triggers
    the Slow Module only upon detection. Concurrently, the Anomaly-Aware Persistent
  Memory (AAPM) proactively safeguards critical visual cues from temporal decay.}
  \label{fig:teaser}
\end{figure*}

Concurrently, the broader field has seen the emergence of general streaming
video understanding frameworks designed to process infinite video streams
\cite{huang2025onlinevideounderstandingovbench, zhou2024streamingdensevideocaptioning,
chen2024videollmonlineonlinevideolarge, xu2025streamingvlmrealtimeunderstandinginfinite}.
By dynamically managing historical context—such as through token merging and persistent
memory \cite{zeng2025streamforestefficientonlinevideo, zhang2024flashvstreammemorybasedrealtimeunderstanding,
qian2024streaminglongvideounderstanding, liu2025streamchatchattingstreamingvideo}
or KV-cache compression \cite{yang2025streammemqueryagnostickvcache, kim2025infinipotvmemoryconstrainedkvcache}—these
models overcome the sequence length limits of offline architectures, enabling continuous
long-term video interaction. However, applying these frameworks directly to the
anomaly domain exposes two critical flaws. First, the transient nature of real-world
anomalies \cite{sultani2018real, du2024uncoveringwhathowcomprehensive} causes their
sparse visual features to be easily diluted during continuous context updates \cite{chen2025aligningeffectivetokensvideo,
zeng2025streamforestefficientonlinevideo}, impairing downstream threat
perception. Second, evaluating every streaming frame with a heavyweight MLLM
incurs prohibitive computational overhead, hindering real-time responsiveness—a
bottleneck that recent decoupled and event-gated architectures have only just begun
to address \cite{qian2025dispiderenablingvideollms, ding2025streammindunlockingframerate,
chatterjee2025memoryefficientstreamingvideollmsrealtime}.

To resolve these intertwined challenges, we propose \textbf{ReactVAU}, a Slow-Fast
Decoupled Framework designed around a reactive event-gated mechanism for
Streaming Video Anomaly Understanding. As illustrated in \Cref{fig:teaser},
the core insight is that a single high-capacity MLLM need not process every
frame uniformly: normal surveillance footage dominates real streams, while
anomalous intervals are rare and short. ReactVAU therefore separates high-frequency
visual anomaly filtering from low-frequency semantic reasoning. A lightweight
Fast Detection Module continuously monitors the stream, an Anomaly-Aware Persistent
Memory (AAPM) preserves transient suspicious evidence while the reasoning model
sleeps, and a heavyweight Slow Reasoning Module is awakened only when the Fast
module detects a potential anomaly. This design keeps the pipeline causal by construction,
reduces redundant MLLM computation on normal content, and still preserves the visual
evidence required for fine-grained causal explanation when a reaction is needed.

The main contributions of our work are summarized as follows:
\begin{itemize}
  \item We propose ReactVAU, a Slow-Fast Decoupled Framework equipped with a
    reactive event-gated mechanism that triggers heavyweight reasoning only
    upon detected anomalies, eliminating the reliance on future frames and enabling
    real-time streaming VAU with reduced computational overhead.

  \item We introduce Spatial Grid Folding (SGF), transforming short-term
    temporal anomaly detection into a highly efficient 2D spatial reasoning
    task that bypasses heavy temporal modeling.

  \item We design the Anomaly-Aware Persistent Memory (AAPM) mechanism,
    utilizing an Anomaly Priority Score and a threat-based Anomaly Pool to
    protect transient abnormal features during continuous memory compression.

  \item Extensive experiments across multiple benchmarks demonstrate that ReactVAU
    achieves competitive performance against state-of-the-art offline models while
    delivering significantly enhanced computational efficiency under a strict causal
    streaming inference protocol.
\end{itemize}

\section{Related Work}
\subsubsection{Video Anomaly Detection.}
Traditional Video Anomaly Detection (VAD) prioritizes event localization via reconstruction-based
one-class classification \cite{lu2013abnormal, hasan2016learning,
gong2019memorizing, reiss2024video} or 3D feature-based Multiple Instance Learning
(MIL) \cite{sultani2018real, wu2020not, tran2015learning, carreira2017quo,
  cao2023newcomprehensivebenchmarksemisupervised, yang2024contextawarevideoanomalydetection,
huang2025trackanomalousobjectgranular}. While foundational, most methods require
offline feature buffering. Pushing towards real-time applications, recent advancements
like REWARD \cite{karim2024real} propose an end-to-end architecture that learns
directly from raw video segments, bypassing offline feature extraction to
achieve highly efficient online detection. Although such low-latency models suit
streaming inputs, they operate strictly as semantic black boxes. Outputting only
scalar probabilities, they fundamentally lack the causal interpretability
essential for comprehensive security analysis.

\subsubsection{Video Anomaly Understanding.}
To bridge this gap, Video Anomaly Understanding (VAU) leverages Multimodal Large
Language Models (MLLMs) for text generation. Initial zero-shot methods
introduced open-world detection~\cite{joo2023cliptsaclipassistedtemporalselfattention,
  ahn2025anyanomalyzeroshotcustomizablevideo,
  liu2026languageguidedopenworldvideoanomaly, lee2025flashbackmemorydrivenzeroshotrealtime,
  chen2025aligningeffectivetokensvideo, shao2025eventvadtrainingfreeeventawarevideo,
zhang2024holmesvadunbiasedexplainablevideo}
by adapting frozen CLIP models (e.g., VadCLIP~\cite{wu2023vadclipadaptingvisionlanguagemodels}),
extracting frame captions for LLMs (LAVAD~\cite{zanella2024harnessinglargelanguagemodels}),
or exploiting anomaly-sensitive attention heads (HeadHunt-VAD~\cite{cai2025headhuntvadhuntingrobustanomalysensitive}).
For deeper reasoning, state-of-the-art models employ supervised instruction-tuning~\cite{du2024uncoveringwhathowcomprehensive,
zhu2025vaur1advancingvideoanomaly, huang2025vadr1videoanomalyreasoning}. For example,
Holmes-VAU~\cite{zhang2025holmesvaulongtermvideoanomaly} utilizes an Anomaly-focused
Temporal Sampler to process untrimmed videos, while VADER~\cite{cheng2025vadercausalvideoanomaly}
employs context-aware sampling and relational encoders to explicitly model causal
object interactions. Additionally, works like VERA~\cite{ye2025veraexplainablevideoanomaly}
integrate verbalized learning, and agentic paradigms such as PANDA~\cite{yang2025pandageneralistvideoanomaly}
and Unified~\cite{lin2025unifiedreasoningframeworkholistic} deploy automated
workflows for holistic causal explanations. Despite their descriptive prowess,
a critical flaw unites these top-tier methods: strict reliance on offline inference.
The necessity to observe entire sequences for global temporal sampling breaks
causality, rendering them fundamentally inapplicable to real-world streaming environments
where future frames remain unobservable.

\subsubsection{Streaming Video Understanding.}
To process infinite video streams, recent works developed streaming frameworks~\cite{huang2025onlinevideounderstandingovbench,
  zhou2024streamingdensevideocaptioning, chen2024videollmonlineonlinevideolarge,
  xu2025streamingvlmrealtimeunderstandinginfinite,
  liu2025streamchatchattingstreamingvideo, kang2025openendedhierarchicalstreamingvideo,
  xiong2025streamingvideounderstandingmultiround, chatterjee2025memoryefficientstreamingvideollmsrealtime,
yan2025learningstreamingvideorepresentation}. They manage unbounded context via
KV-cache compression~\cite{yang2025streammemqueryagnostickvcache, kim2025infinipotvmemoryconstrainedkvcache}
or persistent memory~\cite{zhang2024flashvstreammemorybasedrealtimeunderstanding,
qian2024streaminglongvideounderstanding}. For instance, StreamForest~\cite{zeng2025streamforestefficientonlinevideo}
organizes visual histories into a memory forest. To mitigate latency, Dispider~\cite{qian2025dispiderenablingvideollms}
and StreamMind~\cite{ding2025streammindunlockingframerate} introduce disentangled
perception-reaction and event-gated cognition. Concurrently, MoniTor~\cite{yang2025monitorexploitinglargelanguage}
deploys MLLMs directly for online anomaly detection. While effective, these models
struggle with VAU's practical demands. First, memory compression algorithms heavily
penalize older events to maintain fixed bounds, rapidly diluting sparse
anomalous features amidst overwhelming normal backgrounds. Second, despite
decoupled designs, continuously querying heavyweight MLLMs for frame-level evaluation
creates severe computational bottlenecks, hindering true real-time
responsiveness.

\section{Method}
To bridge the conflicting demands of real-time efficiency and deep causal
explainability in streaming environments, we present \textbf{ReactVAU}, a Slow-Fast
Decoupled Framework. As illustrated in \Cref{fig:architecture}, ReactVAU strictly
operates under a causal streaming protocol, accessing only current and
historical visual information, and is structured around three components:
\textbf{(1) Fast Detection Module} (\cref{sec:SGF}), which continuously
filters streaming frames for anomalies without heavy temporal modeling; \textbf{(2)
Anomaly-Aware Persistent Memory (AAPM)} (\cref{sec:AAPM}), which accumulates visual
context while shielding transient threat features from memory compression; and
\textbf{(3) Slow Reasoning Module} (\cref{sec:SlowReasoning}), a heavyweight
MLLM that remains dormant during normal conditions and awakens only upon
anomaly verification to perform fine-grained causal reasoning on the protected
memory.

\subsection{Preliminary: Continuous Memory Formulation}
To process infinite video streams without catastrophic memory overflow, our framework
explicitly leverages the StreamForest-7B architecture \cite{zeng2025streamforestefficientonlinevideo}
as the backbone for our heavyweight Slow Reasoning Module. Designed
exclusively for streaming comprehension, we formalize its native memory management
mechanism into the following continuous tiers:

\begin{figure*}[t]
  \centering
  \includegraphics[width=\linewidth]{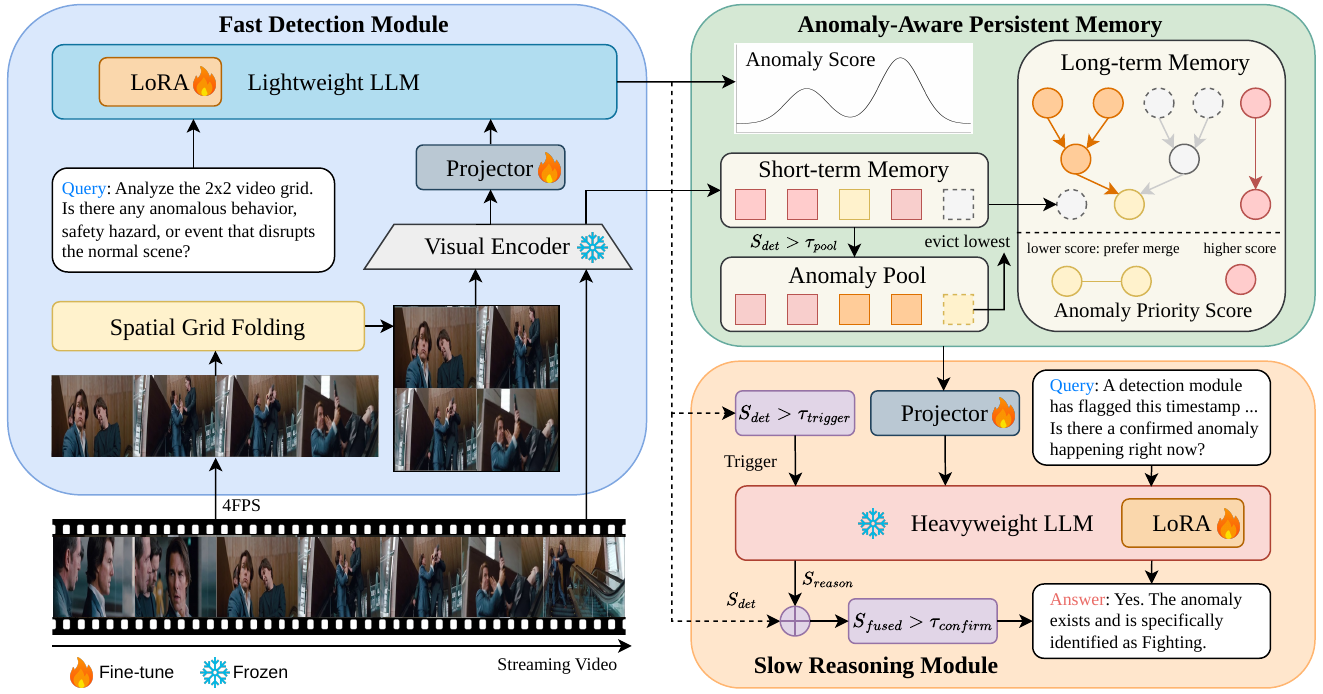}
  \caption{\textbf{Architecture of ReactVAU.} The proposed Slow-Fast Decoupled
    Framework comprises three core components: a lightweight Fast Detection
    module, an Anomaly-Aware Persistent Memory (AAPM), and a heavyweight Slow
    Reasoning Module. The Fast Module continuously processes video streams via Spatial
    Grid Folding (SGF) to yield a real-time anomaly score ($S_{det}$). This
    score directs the AAPM to protect critical anomalous features from
    compression decay. Once an anomaly is detected ($S_{det}> \tau_{trigger}$),
    the dormant Slow Module is awakened to conduct fine-grained causal reasoning
    on the safeguarded memory sequence, generating a secondary verification score
    ($S_{reason}$) and a semantic description. The final detection outcome is
  dictated by the fused score ($S_{fused}$).}
  \label{fig:architecture}
\end{figure*}

\noindent
\textbf{Real-Time Perception (RTP).} RTP preserves a complete set of feature tokens
from the current frame ($N_{RTP}=729$), providing fine-grained spatial
information near the current timestamp.

\noindent
\textbf{Fine-grained Spatiotemporal Window (FSTW).} FSTW stores compressed visual
features of recent frames ($N_{ST}=128$ tokens per frame) within a fixed 12-frame
sliding window. When the window capacity is exceeded, overflowing frames are grouped
into meta-event nodes and evicted to long-term storage.

\noindent
\textbf{Persistent Event Memory Forest (PEMF).} PEMF
\cite{zeng2025streamforestefficientonlinevideo} organizes long-term visual
history as compressed event nodes ($N_{PEMF}=64$ tokens per node) under a
token quota $C_{max}=2048$. When this quota is exceeded, PEMF computes a
composite penalty $\mathcal{P}_{total}$ from similarity, temporal distance,
and merge frequency, and iteratively merges the adjacent node pair with the lowest
penalty. This preserves recent and distinctive events while heavily
compressing redundant or distant history.

\subsection{Fast Detection Module and Spatial Grid Folding (SGF)}
\label{sec:SGF} Existing VAD models often rely on computationally expensive 3D
CNNs \cite{tran2015learning, carreira2017quo} to extract temporal features. While
recent Video LLMs \cite{maaz2024videochatgpt, zhang2023videollama,
lin2024videollava, chen2024internvl2} capture comprehensive temporal sequences,
their sequence-level processing in a streaming context imposes heavy overhead
on computational resources. Inspired by recent findings demonstrating that Vision-Language
Models can perform temporal reasoning spatially when frames are stitched together
\cite{kim2024imagegridworthvideo, doorenbos2025videopanelslongvideo}, we
introduce \textbf{Spatial Grid Folding (SGF)}. This mapping of temporal
dynamics into 2D spatial layouts allows the lightweight VLM to leverage its
inherent spatial attention mechanisms, thereby exploiting the powerful priors acquired
during large-scale pre-training without the computational burden of heavy temporal
modeling.

We utilize PaliGemma2-3B \cite{beyer2024paligemma} as the Fast Detection backbone.
Its visual encoder (SigLIP-So400m \cite{zhai2023sigmoid}) $\mathcal{E}_{vis}$
matches the one used in the heavyweight Slow Reasoning Module, providing consistent
visual extraction priors across the framework. To equip the module with
anomaly perception capabilities tailored for SGF, the model is fine-tuned for 1
epoch on a specifically constructed \textbf{Grid Image Dataset} derived from
UCF-Crime \cite{sultani2018real} and XD-Violence \cite{wu2020not}, updating only
the projector and LoRA \cite{hu2021loralowrankadaptationlarge} weights attached
to the LLM. The detailed construction and balancing strategies of this dataset
are provided in the Supplementary Material.

During streaming inference at 4 FPS, a sliding window of 1 second (frames
$F_{1}$ to $F_{4}$) is spatially folded into a $2 \times 2$ Grid Image,
denoted as $I_{grid}$. This grid is processed by $\mathcal{E}_{vis}$ and mapped
to the language layer via a projector $\Psi_{det}$. Combined with the task
prompt $\mathcal{Q}_{det}$, the lightweight language model ($\text{LLM}_{det}$)
predicts the binary anomaly target logits $z_{\text{"Yes"}}$ and $z_{\text{"No"}}$.
The continuous detection anomaly score $S_{det}$ is derived using a localized
Softmax over these targets:
\begin{equation}
  S_{det}= \frac{\exp(z_{\text{"Yes"}})}{\exp(z_{\text{"Yes"}})+\exp(z_{\text{"No"}})}
  \label{eq:det_score}
\end{equation}
This is mathematically equivalent to a sigmoid over $z_{\text{"Yes"}}-z_{\text{"No"}}$,
but follows the native next-token prediction interface of the language model
and focuses calibration on the decision boundary between the two target tokens.
For a binary grid label $b\in\{0,1\}$, the same localized distribution is
trained with
\begin{equation}
  \mathcal{L}_{det}= -b\log S_{det}- (1-b)\log(1-S_{det}). \label{eq:det_loss}
\end{equation}
The Slow Reasoning Module is awakened only when $S_{det}> \tau_{trigger}$, where
$\tau_{trigger}$ is calibrated from the training split as described in the Supplementary
Material.

\subsection{Anomaly-Aware Persistent Memory (AAPM)}
\label{sec:AAPM} Because the Slow Reasoning Module remains dormant during normal
streams, the visual history must be updated independently of heavyweight reasoning.
Yet generic streaming memory treats dominant normal backgrounds and rare transient
anomalies uniformly, allowing brief anomalous evidence to be compressed into background events
before later retrieval. We introduce \textbf{Anomaly-Aware Persistent Memory (AAPM)}, which
retains the FSTW/PEMF hierarchy while using $S_{det}$ at three complementary
horizons: long-term event protection, high-density evidence retention, and
fine-grained current perception.

\subsubsection{Anomaly Priority Score.}
The native PEMF consolidates adjacent event nodes according to similarity
($P_{s}$), temporal distance ($P_{t}$), and merge frequency ($P_{m}$). Although
effective for generic streaming content, these criteria cannot distinguish a rare anomaly
from ordinary background, so an old but critical event may be repeatedly merged. We introduce an
\textbf{Anomaly Priority Score ($P_{a}$)} that assigns high-threat node pairs an
exponential protective weight:

\begin{equation}
  P_{a}(i, j) = \exp\left(\kappa \cdot \frac{S_{det}^{(i)}+S_{det}^{(j)}}{2}\right
  ) \label{eq:aps}
\end{equation}
where $\kappa$ is a scaling constant and $S_{det}^{(i)}$ is the Fast Detection
score associated with node $i$ when it enters PEMF. This score is added to the
native total penalty:

\begin{equation}
  \mathcal{P}_{total}'(i, j) = w_{s}P_{s}(i, j) + w_{t}P_{t}(i, j) + w_{m}P_{m}
  (i, j) + w_{a}P_{a}(i, j) \label{eq:total_penalty}
\end{equation}
where $w_{s},w_{t},w_{m}$ inherit the native PEMF weights and $w_{a}$ controls
anomaly protection. Because PEMF merges the pair with the lowest penalty,
adding $P_{a}$ makes anomaly-rich pairs less likely to be consolidated. Normal
segments retain the original PEMF behavior, while long-term anomalous events
remain semantically distinguishable despite temporal aging.

\subsubsection{Anomaly Pool.}
The Anomaly Priority Score protects event semantics, but PEMF nodes still undergo
spatial compression and hierarchical merging. Fine visual details needed for
causal verification may therefore be lost even when an event node survives. To
preserve such evidence, we introduce an isolated \textbf{Anomaly Pool
($\mathcal{M}_{pool}$)} that stores higher-density features from suspicious
frames as evidence anchors for the Slow Module. The pool holds eight frames at
128 tokens per frame and lies outside the $C_{max}=2048$ PEMF quota, preventing
anomaly evidence from displacing the normal background required for comparative
reasoning. To avoid contamination by minor score fluctuations, only frames satisfying
$S_{det}>\tau_{pool}$ enter $\mathcal{M}_{pool}$. Once full, the pool retains
the strongest evidence by evicting the frame with the lowest anomaly score:

\begin{equation}
  e_{evict}= \arg\min_{k \in \mathcal{M}_{pool}}S_{det}^{(k)}\label{eq:anomaly_pool}
\end{equation}

\subsubsection{Dynamic Dense Sampling and Real-Time Anomaly Perception.}
Long-term protection alone cannot recover rapid motion that was never encoded at sufficient
temporal resolution. AAPM therefore adapts current perception to the Fast score. During normal intervals ($S_{det}\le\tau_{trigger}$), it
updates memory sparsely at 1 FPS using the last frame of each one-second grid, avoiding redundant processing of static background. Once
$S_{det}>\tau_{trigger}$, it switches to dense perception and independently encodes all four frames as
$\mathcal{M}_{RTP}$, providing the Slow Module with local micro-dynamics of the suspected event alongside
anomaly-protected historical context. The complete capacity allocation of the assembled memory is provided in
Supplementary Material.

\subsection{Slow Reasoning Module}
\label{sec:SlowReasoning} When the trigger threshold ($\tau_{trigger}$) is breached,
the dormant 7B MLLM is awakened and retrieves the fully constructed visual
memory sequence $\mathcal{M}_{seq}$ from the AAPM. This sequence is structured
chronologically and hierarchically:

\begin{equation}
  \mathcal{M}_{seq}= \mathcal{M}_{PEMF}\oplus \mathcal{M}_{ST}\oplus \mathcal{M}
  _{pool}\oplus \mathcal{M}_{RTP}\label{eq:mem_seq}
\end{equation}

where $\mathcal{M}_{ST}$ denotes the short-term component of FSTW and
$\mathcal{M}_{RTP}$ denotes the dense RTP tokens produced by AAPM. The raw visual
tokens comprising $\mathcal{M}_{seq}$ are generated by the shared vision encoder
$\mathcal{E}_{vis}$.

To perform fine-grained secondary verification, the visual memory sequence
$\mathcal{M}_{seq}$ is mapped into the language space via the reasoning
projector $\Psi_{reason}$ and combined with the reasoning prompt
$\mathcal{Q}_{reason}$. The resulting sequence is processed by
$\text{LLM}_{reason}$. Following the Fast Module's scoring formulation
(\cref{eq:det_score}), a secondary anomaly score $S_{reason}$ is computed from
the ``Yes'' and ``No'' logits.

The final system anomaly score $S_{fused}$ is a weighted fusion of the Fast Detection
score ($S_{det}$) and the Slow Reasoning score ($S_{reason}$):

\begin{equation}
  S_{fused}= \mu \cdot S_{det}+ (1 - \mu) \cdot S_{reason}\label{eq:fused_score}
\end{equation}

where $\mu$ is set to 0.4 based on empirical validation. The slightly larger
weight on $S_{reason}$ reflects its deep semantic reasoning over both the current
observation and accumulated history, while $S_{det}$ preserves sensitivity to
abrupt local changes.

To determine the final system action, we establish a confirmation threshold $\tau
_{confirm}$. Only if $S_{fused}> \tau_{confirm}$ is the anomaly verified. Upon
confirmation, the Reasoning Module generates the fine-grained anomaly
description $\mathcal{D}= \{w_{1}, w_{2}, \dots, w_{N}\}$ by maximizing the
autoregressive likelihood conditioned on the accumulated memory sequence and
prompt:

\begin{equation}
  P(\mathcal{D}\mid\mathcal{M}_{seq},\mathcal{Q}_{reason}) = \prod_{n=1}^{N}P(w
  _{n}\mid w_{<n}, \mathcal{M}_{seq}, \mathcal{Q}_{reason}) \label{eq:text_output}
\end{equation}

During instruction tuning, the Slow Reasoning Module uses standard full-vocabulary
teacher-forced next-token cross-entropy:
\begin{equation}
  \mathcal{L}_{reason}= -\sum_{n=1}^{N}\log P(w_{n}\mid w_{<n}, \mathcal{M}_{seq}
  , \mathcal{Q}_{reason}). \label{eq:reason_loss}
\end{equation}

This formulation ensures that the generated narrative is rigorously grounded
in the verified visual evidence comprehensively retrieved from the anomaly-protected
memory structures.

\section{Experiments}
\label{sec:blind}

\subsection{Benchmark Datasets and Evaluation Metrics}
To rigorously evaluate the performance of ReactVAU, we conduct experiments
across three dataset benchmarks that encompass VAD and VAU tasks:

\noindent
\textbf{UCF-Crime \cite{sultani2018real}:} Comprises 1,900 untrimmed surveillance
videos covering 13 anomaly categories. We report the frame-level Area Under
the Receiver Operating Characteristic Curve (AUC).

\noindent
\textbf{XD-Violence \cite{wu2020not}:} Contains 4,754 untrimmed videos focusing
exclusively on violent events. Due to extreme class imbalance, we report the Average
Precision (AP) alongside the AUC.

\noindent
\textbf{HIVAU-70K \cite{zhang2025holmesvaulongtermvideoanomaly}:} A
comprehensive VAU benchmark providing over 70,000 annotations at Clip, Event,
and Video granularities. To evaluate semantic explanations, we utilize
standard generation metrics: BLEU \cite{papineni2002bleu}, METEOR
\cite{banerjee2005meteor}, ROUGE-L \cite{lin2004rouge}, and primarily CIDEr
\cite{vedantam2015cider}, which effectively prioritizes critical rare anomaly descriptors
over generic text.

Detailed evaluation protocols, including causal online smoothing and frame-level
score broadcasting mechanisms designed to adapt VAD/VAU tasks into streaming
constraints, are provided in the Supplementary Material.

\subsection{Experimental Results}
\subsubsection{Video Anomaly Detection (VAD) Performance.}
We compare ReactVAU
against state-of-the-art VAD methods, categorizing them by their operational settings:
weakly supervised, offline fine-tuned, offline training-free, and online architectures.

\begin{table}[tb]
  \caption{\textbf{Video Anomaly Detection Performance.} Comparison on UCF-Crime
    and XD-Violence datasets, categorized by inference paradigms and training
    strategies. StreamForest$^{\dagger}$ is fine-tuned on the HIVAU instruction
  dataset.}
  \label{tab:vad_performance}
  \centering
  \footnotesize
  \setlength{\tabcolsep}{5pt}
  \begin{tabular}{@{} ll c cc @{}}
    \toprule \multirow{2}{*}{Setting}                        & \multirow{2}{*}{Method}                                             & UCF-Crime      & \multicolumn{2}{c}{XD-Violence} \\
    \cmidrule(lr){3-3} \cmidrule(lr){4-5}                    &                                                                     & AUC (\%)       & AP (\%)                        & AUC (\%)       \\
    \midrule \multirow{2}{*}{\textit{Offline Weakly Sup.}}   & CLIP-TSA~\cite{joo2023cliptsaclipassistedtemporalselfattention}     & 87.58          & 82.19                          & -              \\
    & VadCLIP~\cite{wu2023vadclipadaptingvisionlanguagemodels}            & \textbf{88.02} & \textbf{84.51}                 & -              \\
    \midrule \multirow{2}{*}{\textit{Offline Fine-Tuned}}    & Holmes-VAD~\cite{zhang2024holmesvadunbiasedexplainablevideo}        & \textbf{89.51} & \textbf{90.67}                 & -              \\
    & Holmes-VAU~\cite{zhang2025holmesvaulongtermvideoanomaly}            & 88.96          & 87.68                          & -              \\
    \midrule \multirow{5}{*}{\textit{Offline Training-Free}} & LLaVA-1.5~\cite{liu2023improved}                                    & 72.84          & 50.26                          & 79.62          \\
    & LAVAD~\cite{zanella2024harnessinglargelanguagemodels}               & 80.28          & 62.01                          & 85.36          \\
    & Unified~\cite{lin2025unifiedreasoningframeworkholistic}             & 84.28          & 68.07                          & \textbf{91.34} \\
    & VERA~\cite{ye2025veraexplainablevideoanomaly}                       & 86.55          & 70.54                          & 88.26          \\
    & HeadHunt-VAD~\cite{cai2025headhuntvadhuntingrobustanomalysensitive} & \textbf{87.03} & \textbf{82.63}                 & -              \\
    \midrule \multirow{2}{*}{\textit{Online Training-Free}}  & Online-LAVAD~\cite{zanella2024harnessinglargelanguagemodels}        & 76.06          & 52.63                          & 76.01          \\
    & MoniTor~\cite{yang2025monitorexploitinglargelanguage}               & \textbf{82.57} & \textbf{55.01}                 & \textbf{79.11} \\
    \midrule \multirow{3}{*}{\textit{Online Fine-Tuned}}     & REWARD~\cite{karim2024real}                                         & 86.94          & 77.71                          & -              \\
    & StreamForest$^{\dagger}$~\cite{zeng2025streamforestefficientonlinevideo}        & 85.26          & 75.92                          & 92.82          \\
    & \textbf{ReactVAU (Ours)}                                            & \textbf{88.44} & \textbf{88.50}                 & \textbf{95.25} \\
    \bottomrule
  \end{tabular}
\end{table}

As presented in \cref{tab:vad_performance}, traditional offline models benefit
immensely from observing the entire video sequence, which inherently inflates
their performance by leveraging future context. Notably, when existing robust
offline models are forced into online settings (e.g., Online-LAVAD~\cite{zanella2024harnessinglargelanguagemodels}), their performance
degrades severely, dropping to 76.06\% AUC on UCF-Crime and 52.63\% AP on XD-Violence.

In stark contrast, operating under strict streaming constraints without any future information, ReactVAU reaches 88.44\% AUC on UCF-Crime
and 88.50\% AP with 95.25\% AUC on XD-Violence. It outperforms existing online
methods, such as MoniTor~\cite{yang2025monitorexploitinglargelanguage} and the closest fine-tuned StreamForest$^{\dagger}$~\cite{zeng2025streamforestefficientonlinevideo} baseline.
Furthermore, our streaming framework remains competitive with the best offline fine-tuned methods (e.g., Holmes-VAD~\cite{zhang2024holmesvadunbiasedexplainablevideo}) that access future frames. These results
show that SGF and Slow-Fast verification can preserve strong anomaly perception
without breaking causality.

\subsubsection{Video Anomaly Understanding (VAU) Performance.}
To evaluate semantic
explanation capabilities, we benchmark ReactVAU on the multi-granular HIVAU-70K
dataset against generic MLLMs and specialized VAU models.
\cref{tab:vau_performance} details the natural language generation metrics
across Clip (C), Event (E), and Video (V) levels. We specifically emphasize the
CIDEr metric in this analysis because it rigorously assigns higher weights to rare,
informative tokens (e.g., specific anomaly behaviors) rather than frequently
occurring background words, making it the most critical indicator of exact anomaly
understanding.

\begin{table}[tb]
  \caption{\textbf{Video Anomaly Understanding Performance on HIVAU-70K.} Methods
    are evaluated across three temporal granularities: clip (C), event (E), and
    video (V). Models denoted with $^{\dagger}$ are fine-tuned on the HIVAU instruction
  dataset.}
  \label{tab:vau_performance}
  \centering
  \setlength{\tabcolsep}{3pt}
  \resizebox{\textwidth}{!}{
    \begin{tabular}{@{} l @{\hspace{8pt}} ccc @{\hspace{10pt}} ccc @{\hspace{10pt}}
      ccc @{\hspace{10pt}} ccc @{}}
      \toprule \multirow{2}{*}{Method}                                                         & \multicolumn{3}{c}{BLEU} & \multicolumn{3}{c}{CIDEr} & \multicolumn{3}{c}{METEOR} & \multicolumn{3}{c}{ROUGE} \\
      \cmidrule(lr){2-4} \cmidrule(lr){5-7} \cmidrule(lr){8-10} \cmidrule(lr){11-13}           & C                        & E                         & V                          & C                        & E              & V              & C              & E              & V              & C              & E              & V              \\
      \midrule \rowcolor[gray]{0.92} \multicolumn{13}{@{}l}{\textit{Generic MLLMs (Zero-shot)}} \\
      Video-ChatGPT~\cite{maaz2024videochatgpt}                                                & 0.152                    & 0.068                     & 0.066                      & 0.033                    & 0.011          & 0.013          & 0.102          & 0.069          & 0.044          & 0.153          & 0.048          & 0.079          \\
      Video-LLAMA~\cite{zhang2023videollama}                                                   & 0.151                    & 0.079                     & 0.104                      & 0.024                    & 0.014          & 0.017          & 0.112          & 0.076          & 0.057          & 0.156          & 0.067          & 0.090          \\
      Video-LLAVA~\cite{lin2024videollava}                                                     & 0.164                    & 0.046                     & 0.055                      & 0.032                    & 0.009          & 0.013          & 0.097          & 0.022          & 0.014          & 0.132          & 0.023          & 0.045          \\
      LLAVA-Next-Video~\cite{zhang2024llavanextvideo}                                          & 0.435                    & 0.091                     & 0.120                      & 0.102                    & 0.015          & 0.031          & 0.117          & 0.085          & 0.096          & 0.198          & 0.080          & 0.106          \\
      QwenVL2~\cite{wang2024qwenvl2}                                                           & 0.312                    & 0.082                     & 0.155                      & 0.044                    & 0.020          & 0.044          & 0.133          & 0.092          & 0.112          & 0.163          & 0.081          & 0.137          \\
      InternVL2~\cite{chen2024internvl2}                                                       & 0.331                    & 0.101                     & 0.145                      & 0.052                    & 0.022          & 0.035          & 0.141          & 0.095          & 0.101          & 0.182          & 0.102          & 0.122          \\
      NVILA~\cite{liu2024nvila}                                                                & 0.610                    & 0.340                     & 0.283                      & 0.261                    & 0.154          & 0.098          & 0.157          & 0.096          & 0.072          & 0.273          & 0.218          & 0.198          \\
      \midrule \rowcolor[gray]{0.92} \multicolumn{13}{@{}l}{\textit{Offline VAU Models}}        \\
      Holmes-VAU$^{\dagger}$~\cite{zhang2025holmesvaulongtermvideoanomaly}                     & 0.913                    & 0.804                     & 0.566                      & 0.467                    & 1.519          & 1.437          & 0.190          & 0.165          & 0.121          & 0.329          & 0.370          & 0.355          \\
      VADER$^{\dagger}$~\cite{cheng2025vadercausalvideoanomaly}                                & \textbf{1.266}           & 1.246                     & 1.268                      & \textbf{1.040}           & 1.763          & 1.812          & \textbf{0.247} & 0.216          & 0.164          & \textbf{0.429} & 0.463          & 0.446          \\
      \midrule \rowcolor[gray]{0.92} \multicolumn{13}{@{}l}{\textit{Streaming VAU Models}}      \\
      StreamForest~\cite{zeng2025streamforestefficientonlinevideo} (Zero-shot)                 & 0.422                    & 0.252                     & 0.271                      & 0.138                    & 0.076          & 0.070          & 0.128          & 0.097          & 0.094          & 0.233          & 0.175          & 0.180          \\
      StreamForest$^{\dagger}$~\cite{zeng2025streamforestefficientonlinevideo}                 & 1.245                    & 1.334                     & 1.320                      & 0.947                    & 1.999          & 1.931          & 0.243          & 0.221          & 0.178          & 0.427          & 0.483          & 0.456          \\
      \textbf{ReactVAU$^{\dagger}$ (Ours)}                                                     & 1.150                    & \textbf{1.366}            & \textbf{1.337}             & 0.920                    & \textbf{2.032} & \textbf{2.016} & 0.222          & \textbf{0.222} & \textbf{0.179} & 0.401          & \textbf{0.488} & \textbf{0.465} \\
      \bottomrule
    \end{tabular}%
  }
\end{table}

As the data shows, generic multi-modal LLMs (e.g., Video-LLAMA~\cite{zhang2023videollama}, QwenVL2~\cite{wang2024qwenvl2},
InternVL2~\cite{chen2024internvl2}) degrade sharply at the Event and Video levels (e.g., InternVL2 scores 0.022 on CIDEr-E
), reflecting the difficulty of retaining transient anomaly
evidence over long contexts. Conversely, at the Clip level, offline VADER performs best because
its global sampler first observes the complete input and concentrates a fixed frame
budget on anomaly-dense regions. ReactVAU instead ingests causally with one AAPM
update per second, leaving short clips with less accumulated evidence. As the Event
and Video contexts grow, AAPM preserves sufficient anomaly evidence for ReactVAU
to achieve the best scores across all reported long-range metrics.

Crucially, the comparison between the native StreamForest and our ReactVAU framework
under the same instruction-tuning setting highlights the impact of temporal
feature dilution in general streaming architectures.
ReactVAU consistently improves Event- and Video-level VAU. These gains support
AAPM's role in preserving critical anomaly evidence from aggressive memory compression.
This preservation ensures that the Slow Reasoning Module has access to anomaly-focused evidence and long-range context,
enabling it to generate highly accurate, causally grounded narratives
regardless of the video's total streaming duration.

\subsubsection{Effectiveness of Slow-Fast Decoupling (Efficiency Analysis).}
A critical bottleneck in streaming Video Anomaly Understanding is the
prohibitive computational cost of evaluating every frame with a heavyweight MLLM. \cref{tab:efficiency_analysis}
reports profiling on a single NVIDIA H100-80GB GPU, demonstrating how our Slow-Fast decoupling fundamentally resolves
this issue. Evaluated on the UCF-Crime test set—where anomalies
naturally occur in approximately 43\% of the segments, ReactVAU successfully
leverages the lightweight Fast Module to filter redundant nominal backgrounds.
This reduces heavyweight 7B-parameter LLM inference queries from 34,670 to
15,955, achieving a 54.0\% reduction in computational workload.

More importantly, this architectural decoupling directly translates to real-time
responsiveness. While the baseline StreamForest (which operates entirely on a
heavy 7B MLLM backbone) suffers from a constant 216.1 ms/query delay, ReactVAU
efficiently restricts the heavyweight LLM to event verification, operating at a mere
22.1 ms/query during normal states. Consequently, the empirical weighted average
streaming latency on UCF-Crime drops to 98.3 ms/query, cutting the total delay
by more than half. Supplementary Material further profiles normal and triggered
paths on an RTX 4090-24GB under a one-second streaming budget to better demonstrate the feasibility of real-world deployment.

\begin{table}[tb]
  \caption{\textbf{Efficiency and Scalability Analysis.} Evaluated on the UCF-Crime
    test set using a single NVIDIA H100-80GB GPU. By employing the lightweight Fast Module as an active gatekeeper, ReactVAU
    significantly minimizes the redundant invocations of the heavyweight Slow Module.
    The weighted average latency is calculated as
    $L_{avg}= L_{normal}\times (1 - \text{Rate}_{anomaly}) + L_{anomaly}\times \text{Rate}
    _{anomaly}$, illustrating theoretical performance under realistic sparse anomaly
  occurrence rates~\cite{sultani2018real}.}
  \label{tab:efficiency_analysis}
  \centering
  \scriptsize
  \setlength{\tabcolsep}{3pt}
  \begin{tabular}{@{} l cc ccc @{}}
    \toprule \multirow{2}{*}{Method / Scenario}                                                                                      & \multicolumn{2}{c}{7B LLM Overhead} & \multicolumn{3}{c}{Latency (ms/query)} \\
    \cmidrule(lr){2-3} \cmidrule(l){4-6}                                                                                             & Queries $\downarrow$                & Reduction                             & Normal & Anomaly & Weighted Avg. $\downarrow$ \\
    \midrule \rowcolor[gray]{0.92} \multicolumn{6}{@{}l}{\textit{Empirical Evaluation (UCF-Crime Test Set, $\sim$43\% Anomaly Rate)}} \\
    StreamForest~\cite{zeng2025streamforestefficientonlinevideo} (Slow-only)                                                                                             & 34,670                              & -                                     & 216.1  & 216.1   & 216.1                      \\
    ReactVAU (Ours)                                                                                                                  & 15,955                              & 54.0\%                                & 22.1   & 269.5   & 98.3                       \\
    \midrule \rowcolor[gray]{0.92} \multicolumn{6}{@{}l}{\textit{Theoretical Projection (Real-World Surveillance Deployment)}}        \\
    ReactVAU (10\% Anomaly Rate)                                                                                                     & $\sim$3,467                         & 90.0\%                                & 22.1   & 269.5   & 39.9                       \\
    ReactVAU (5\% Anomaly Rate)                                                                                                      & $\sim$1,734                         & \textbf{95.0\%}                       & 22.1   & 269.5   & \textbf{31.0}              \\
    \bottomrule
  \end{tabular}
\end{table}

\noindent
\textbf{Real-world Scalability Projection:} It is crucial to note that the UCF-Crime
test set is highly condensed with anomalous events. In actual real-world
surveillance deployment, true anomalies are exceedingly rare and typically account
for less than 5\% to 10\% of the continuous video stream~\cite{sultani2018real}.
As projected in the bottom section of \cref{tab:efficiency_analysis}, under a realistic
5\% anomaly rate, the trigger frequency of our Slow Module plummets. This
theoretical projection indicates that ReactVAU could reduce 7B LLM
queries by up to 95.0\%, driving the average operational latency down to
31.0 ms/query. This confirms that ReactVAU is not only empirically
effective but highly scalable for infinite real-world streams.

\subsection{Qualitative Results}
\label{sec:qualitative}

Evaluating comprehensive anomaly events requires understanding at multiple semantic
granularities, from localized clip-level actions to holistic video-level summaries.
Conventional offline VAU models typically adapt global temporal sampling to each
specific question. This necessitates separate full-video processing passes for every query,
which incurs massive redundant computation and fundamentally breaks causality.

\begin{figure*}[t]
  \centering
  \includegraphics[width=\linewidth]{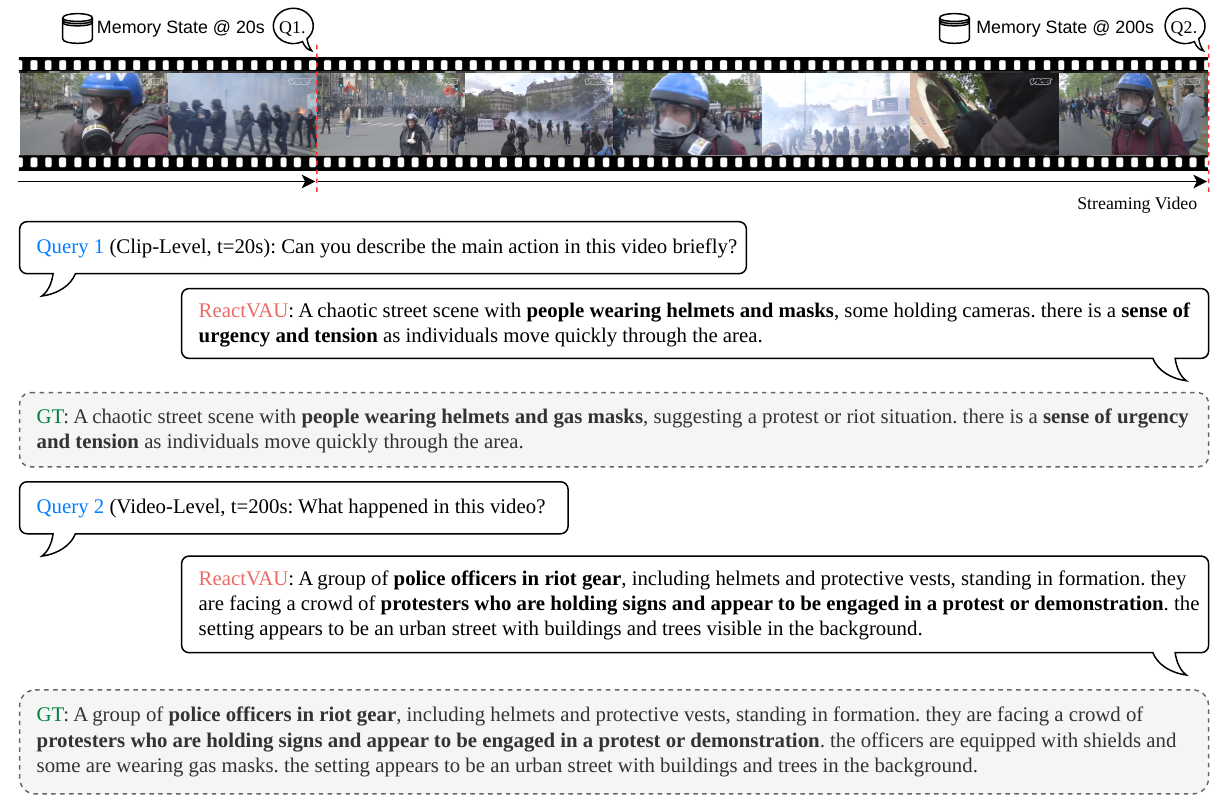}
  \caption{\textbf{Qualitative streaming VAU result.} ReactVAU preserves anomaly
  evidence in AAPM while processing the stream causally. Each query is answered
  using the memory state available at its corresponding timestamp, e.g., $0$--$20$s
  memory for a clip-level query at $t=20$s and $0$--$200$s memory for a video-level
  summary at $t=200$s, without question-specific global resampling.}
  \label{fig:qualitative}
\end{figure*}

In contrast, ReactVAU performs VAU within a causal streaming pipeline. As illustrated
in \cref{fig:qualitative}, ReactVAU processes the stream sequentially without
observing future frames, while the AAPM protects
crucial anomalous evidence (e.g., escaping rioters and police formations) from being progressively diluted
within normal streaming context. The resulting continuously updated memory can be queried at the
corresponding timestamp, such as a clip-level query at $t=20$s or a video-level
summary at $t=200$s, allowing ReactVAU to capture both transient
localized actions and the global event narrative without resampling or re-encoding
the full video.

\subsection{Ablation Study}
To rigorously validate the effectiveness of our proposed modules, we conduct
an incremental ablation study (\cref{tab:ablation}). Given the highly imbalanced
nature of real-world anomalies, we primarily focus on the Average Precision (AP)
metric on XD-Violence, which strictly evaluates a model's ability to suppress false
positives—a critical capability enhanced by our framework.

\begin{table}[t]
  \centering
  \small
  \caption{\textbf{Ablation Study.} Incremental validation of key framework components.
    \textbf{PaliGemma2}: PaliGemma2-3B, \textbf{StreamForest}: StreamForest-7B, \textbf{SGF}: Spatial
    Grid Folding. The strictly causal online smoothing mechanism is uniformly applied
  across all streaming evaluations.}
  \label{tab:ablation}
  \begin{tabular}{@{}lccc@{}}
    \toprule \multirow{2}{*}{\textbf{Method / Incremental Component}}                                       & \textbf{UCF-Crime} & \multicolumn{2}{c}{\textbf{XD-Violence}} \\
    \cmidrule(lr){2-2} \cmidrule(l){3-4}                                                                    & AUC                & AP                                      & AUC            \\
    \midrule \rowcolor[gray]{0.92} \multicolumn{4}{@{}l}{\textit{Fast Module Analysis (PaliGemma2 Base)}}            \\
    Single Frame (Zero-shot)                                                                                & 74.42              & 53.71                                   & 79.71          \\
    SGF (Zero-shot)                                                                                         & 69.45              & 43.42                                   & 75.44          \\
    SGF (Fine-tuned)                                                                                        & 85.37              & 79.55                                   & 91.18          \\
    \midrule \rowcolor[gray]{0.92} \multicolumn{4}{@{}l}{\textit{ReactVAU Architecture Evolution (StreamForest Base)}} \\
    Baseline (Zero-shot)                                                                                    & 78.09              & 49.24                                   & 81.94          \\
    Baseline (Fine-tuned)                                                                                   & 85.26              & 75.92                                   & 92.82          \\
    \quad + Slow-Fast Decoupling (Fast Tuned)                                                               & 85.66              & 80.59                                   & 91.34          \\
    \quad + Slow-Fast Decoupling (Both Tuned)                                                               & 87.17              & 86.90                                   & 94.16          \\
    \quad \textbf{+ AAPM (Ours)}                                                                            & \textbf{88.44}     & \textbf{88.50}                          & \textbf{95.25} \\
    \bottomrule
  \end{tabular}
\end{table}

\noindent
\textbf{Effectiveness of Spatial Grid Folding (SGF):} The upper block isolates
the Fast Module. Transitioning PaliGemma2-3B from single-frame
inputs to zero-shot grid images disrupts its pre-trained spatial priors, causing
the AP to drop (53.71\%→43.42\%). However, upon fine-tuning on our Grid Image Dataset,
the SGF mechanism achieves a massive AP surge to 79.55\%. This validates that
properly adapted Vision-Language Models can perform robust temporal anomaly
perception within a folded 2D layout.

\noindent
\textbf{Evolution from Streaming Baseline to ReactVAU:} The lower block contrasts
the continuous memory baseline against our architectural evolution. The zero-shot
StreamForest struggles severely with anomaly semantics (49.24\% AP), though
domain fine-tuning elevates it to 75.92\% AP. Building upon this, introducing our
Slow-Fast Decoupled Framework yields immediate gains. By leveraging
the Fast Module (with SGF) as a rapid filter and the Slow Module as a rigorous
semantic gatekeeper, this dual-verification mechanism effectively eradicates
false alarms, driving the AP up to 86.90\% when both modules are tuned.

Crucially, integrating the AAPM allows the lightweight Fast Module to directly
govern evidence retention, effectively safeguarding transient abnormal features.
This integration pushes the streaming performance to a highly
competitive 88.50\% AP (95.25\% AUC) on XD-Violence and 88.44\% AUC on UCF-Crime.
These consistent improvements across anomaly benchmarks underscore
the architectural synergy between decoupled verification and anomaly-protected memory,
highlighting that both are structurally important for infinite stream understanding.

\begin{figure*}[tb]
  \centering
  \begin{minipage}[t]{0.48\textwidth}
    \vspace{0pt}
    \centering
    \includegraphics[width=\textwidth]{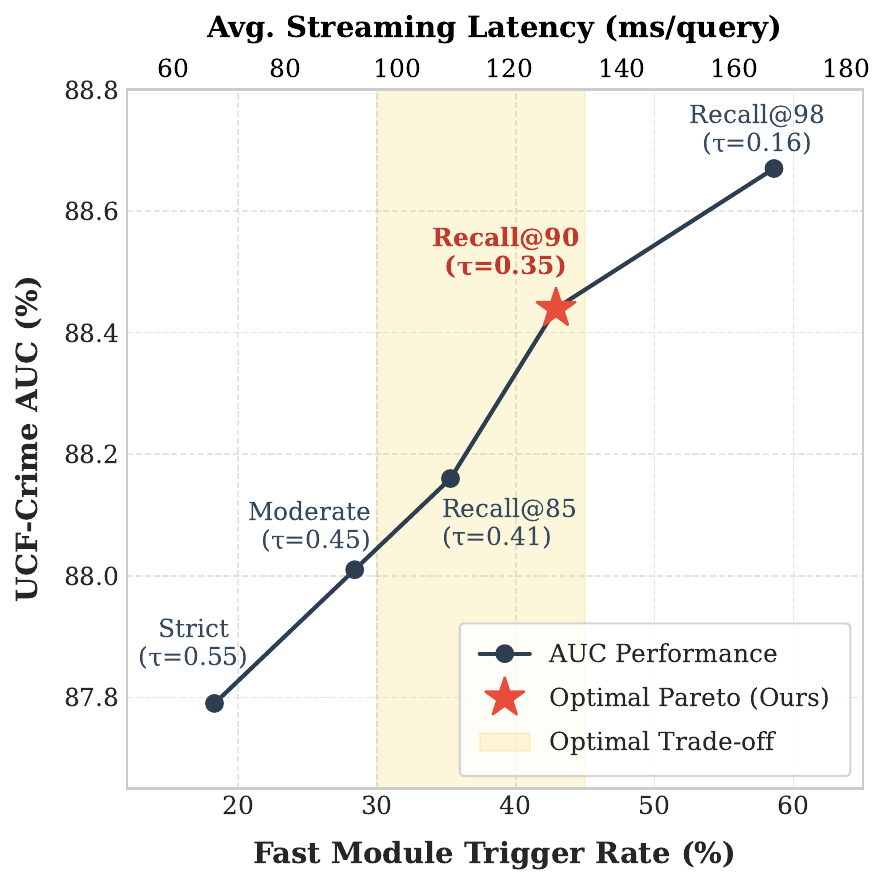}
    \caption{\textbf{Efficiency vs. Accuracy.} Pareto trade-off analysis on UCF-Crime
    demonstrating the optimization of the trigger threshold.}
    \label{fig:threshold_tradeoff}
  \end{minipage}
  \hfill
  \begin{minipage}[t]{0.48\textwidth}
    \vspace{0pt}
    \centering
    \captionof{table}{\textbf{Score Fusion Strategies.} Ablation on late fusion.
      For compact formulas, $D$, $R$, and $F$ denote the full paper notation
      $S_{det}$, $S_{reason}$, and $S_{fused}$, respectively. Purely replacing the fused score
      with the Slow score (Replace) fails to capture temporal dynamics, while Weighted fusion combines
    the spatial sensitivity of Fast Detection and semantic verification of Slow Reasoning.}
    \label{tab:fusion_strategy}
    \vspace{6pt}
    \resizebox{\textwidth}{!}{
      \begin{tabular}{@{}lc@{}}
        \toprule \textbf{Strategy (Formulation)}             & \textbf{AUC (\%)} \\
        \midrule Replace ($F=R$)                       & 87.28             \\
        Adaptive ($F=(1-R)D+R^{2}$)     & 88.08             \\
        \midrule Weighted ($F=0.3D+0.7R$)       & 88.17             \\
        \textbf{Weighted ($\mathbf{F=0.4D+0.6R}$)} & \textbf{88.44} \\
        Weighted ($F=0.5D+0.5R$)                & 88.26             \\
        \bottomrule
      \end{tabular}
    }
  \end{minipage}
\end{figure*}

\noindent
\textbf{Threshold Trade-off:} \Cref{fig:threshold_tradeoff} illustrates the Pareto
front between computational cost and accuracy by modulating the trigger threshold
($\tau_{trigger}$). While lower thresholds increase sensitivity at the cost of
higher overhead, higher thresholds reduce latency but risk missing subtle anomalies.
A well-calibrated threshold successfully strikes an optimal balance, effectively
filtering out the vast majority of redundant nominal frames while ensuring that
critical threat deviations are not overlooked. This optimization significantly
enhances the overall system utility and demonstrates the extreme flexibility
of ReactVAU in adapting to varying hardware constraints. Cross-dataset threshold
transfer is reported in Supplementary Material.

\noindent
\textbf{Score Fusion Strategy:} \Cref{tab:fusion_strategy} evaluates
the fusion of the Fast ($S_{det}$) and Slow ($S_{reason}$) anomaly scores. Our empirical
results reveal that purely overriding the Fast score with the reasoning MLLM's
prediction (Replace strategy) yields suboptimal results, dropping the AUC to
87.28\%. Although the Slow Module receives complete dense frames upon
activation, its objective heavily prioritizes long-form semantic reasoning over
the entire historical sequence. This broad contextual focus inadvertently
smooths out instantaneous high-frequency threat spikes. Conversely, the Fast
Module is exclusively calibrated for short-term spatial perception, maintaining
extreme sensitivity to sudden visual disruptions. Consequently, an adaptive fusion
approach falls short of optimal synergy, whereas a complementary weighted
fusion ($S_{fused}=0.4S_{det}+0.6S_{reason}$) optimally combines localized reactive

\section{Conclusion}
In this paper, we introduced ReactVAU, a Slow-Fast Decoupled Framework for streaming
Video Anomaly Understanding that reconciles strict causal access, low-latency
operation, and high-capacity semantic reasoning. ReactVAU continuously monitors
incoming streams with a lightweight Fast Detection Module built on Spatial Grid
Folding, preserves transient threat evidence through Anomaly-Aware Persistent
Memory, and reactively awakens a heavyweight Slow Reasoning Module only when
semantic verification is needed. Extensive evaluations on UCF-Crime, XD-Violence,
and HIVAU-70K demonstrate that ReactVAU remains competitive with offline methods
that can observe future frames while substantially reducing redundant heavyweight
MLLM invocations. Beyond improving accuracy alone, ReactVAU improves the
accuracy--efficiency Pareto frontier for causal streaming VAU, offering a practical
path toward deploying semantic anomaly reasoning in continuous real-world
monitoring systems.

\section*{Acknowledgements}
This work is supported by the NVIDIA Taiwan AI Research \& Development Center
(TRDC).

%
%
\bibliographystyle{splncs04}
\bibliography{reference}
\end{document}